\documentclass[sigconf]{acmart}
\usepackage{booktabs} 
\usepackage{multirow}
\usepackage{graphicx}
\usepackage{subfigure}
\usepackage{algorithm}
\usepackage{algorithmic}
\usepackage{color}
\usepackage{amsmath}
\usepackage{textcomp}
\usepackage{bm}
\usepackage{hyperref}
\usepackage{balance}
\usepackage[T1]{fontenc}
\setcopyright{iw3c2w3}
\copyrightyear{2021}
\acmYear{2021}
\acmConference[WWW '21]{Proceedings of the Web Conference 2021}{April 19--23, 2021}{Ljubljana, Slovenia}
\acmBooktitle{Proceedings of the Web Conference 2021 (WWW '21), April 19--23, 2021, Ljubljana, Slovenia}
\acmPrice{}
\acmDOI{10.1145/3442381.3449828}
\acmISBN{978-1-4503-8312-7/21/04}
\renewcommand{\shortauthors}{}

\begin{document}

\title{Improving Cyberbullying Detection with User Interaction}

\begin{abstract}
Cyberbullying, identified as intended and repeated online bullying behavior, has become increasingly prevalent in the past few decades. 
Despite the significant progress made thus far, the focus of most existing work on cyberbullying detection lies in the independent content analysis of different comments within a social media session.
We argue that such leading notions of analysis suffer from three key limitations: they overlook the temporal correlations among different comments;
they only consider the content within a single comment rather than the topic coherence across comments; 
they remain generic and exploit limited interactions between social media users.
In this work, we observe that user comments in the same session may be inherently related, e.g., discussing similar topics, and their interaction may evolve over time. 
We also show that modeling such topic coherence and temporal interaction are critical to capture the repetitive characteristics of bullying behavior, thus leading to better predicting performance. 
To achieve the goal, we first construct a unified temporal graph for each social media session. 
Drawing on recent advances in graph neural network, we then propose a principled graph-based approach for modeling the temporal dynamics and topic coherence throughout user interactions. 
We empirically evaluate the effectiveness of our approach with the tasks of session-level bullying detection and comment-level case study.
Our code is released to public.~\footnote{https://github.com/gesy17/TGBully}
\end{abstract}



\author{Suyu Ge$^1$, Lu Cheng$^2$, Huan Liu$^2$}
\affiliation{%
  \institution{$^1$Department of Electronic Engineering, Tsinghua University\\
  $^2$School of Computer Science and Engineering, Arizona State University\\
  gesy17@mails.tsinghua.edu.cn, \{lcheng35,huanliu\}@asu.edu}
}


\maketitle
\section{Introduction}
% Identified as a form of bullying or harassment using electronic means, cyberbullying is characterized by posting rumors, threats, or pejorative labels on social media~\cite{us_legal}.
% It has become increasingly prevailing with the expansion of digital sphere and advancement of technology. 
% According to the 2019 Youth Risk Behavior Surveillance System (Centers for Disease Control and Prevention), an estimated 15.7\% of high school students were electronically bullied in the 12 months prior to the survey~\cite{centersYRBSS}.
% In response to the rapid growing number of reported cases, numerous efforts have been put into automatic detection and mitigation of cyberbullying behaviors.
% Both psychology analysis~\cite{elsherief2018hate,dooley2009cyberbullying} and computational modeling~\cite{wu2010modeling,zhao2016cyberbullying} have been exploited to tackle the socio-technical problem.
Cyberbullying is identified as a form of bullying or harassment through electronic means, which is typically characterized by posting rumors, threats, or pejorative labels on social media~\cite{us_legal}. 
Due in part to the low cost and easy access to social networking sites, Cyberbullying has become increasingly prevalent in recent decades.
According to the 2019 Youth Risk Behavior Surveillance System (Centers for Disease Control and Prevention), an estimated 15.7\% of high school students were electronically bullied in the 12 months prior to the survey~\cite{centersYRBSS}. 
In response to the rapidly growing number of cyberbullying cases, numerous efforts (e.g., \cite{wu2010modeling,zhao2016cyberbullying,dani2017sentiment}) have been put into computational detection and mitigation of cyberbullying behaviors to tackle the socio-technical problem.
\begin{figure}[t]
	\centering
	\resizebox{0.8\columnwidth}{!}{\includegraphics{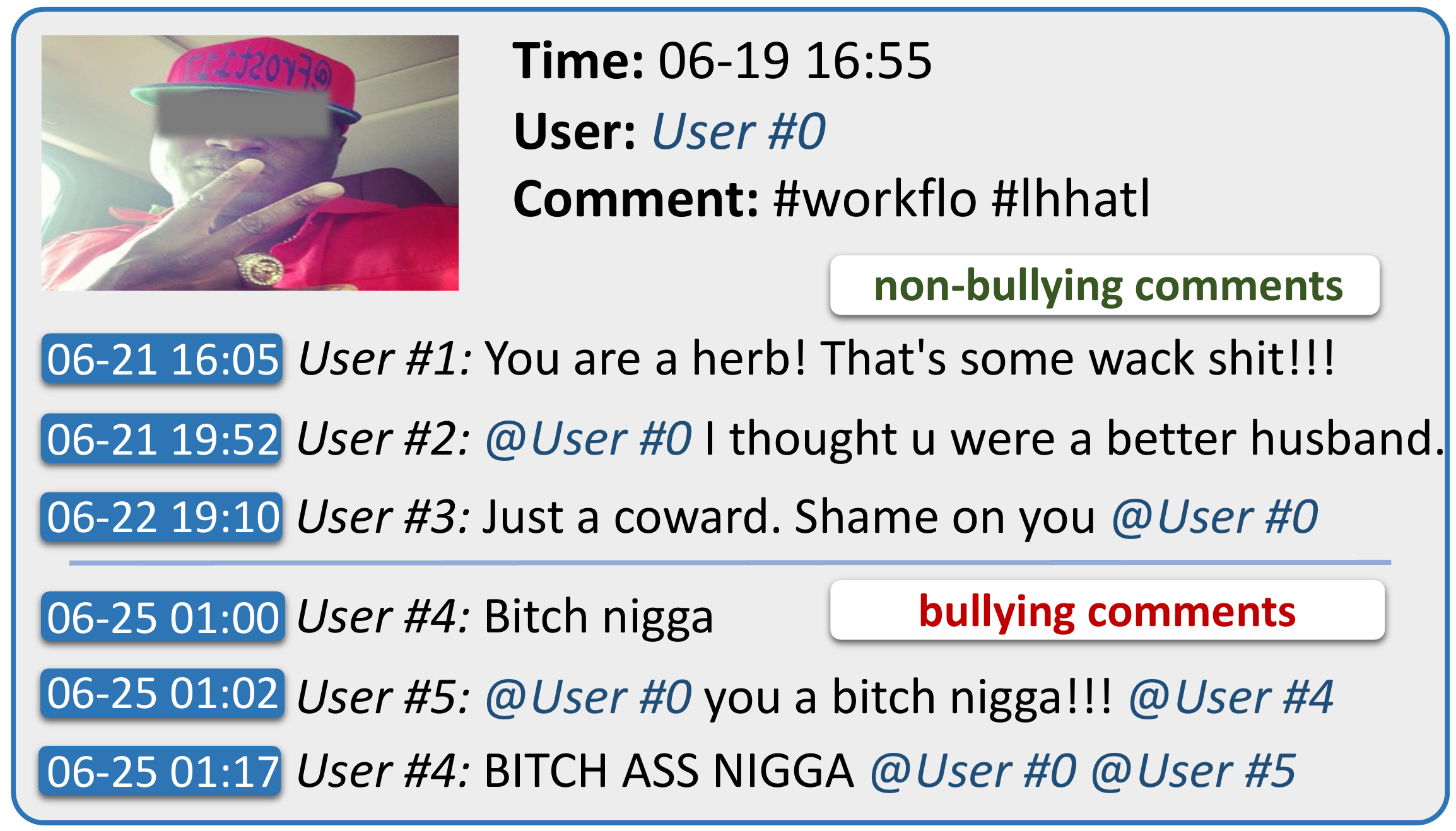}}
	\caption{An example of cyberbullying session consisting of the original post, user comments and posted time. 
	The three comments in the upper part are non-bullying comments while those in the lower part are regarded as bullying comments.}
	\label{fig:intro}
\end{figure}

Despite the remarkable progress made towards computational cyberbullying detection, we argue that it is still essential to understand how users in a social media session interact with each other in order to mitigate the long-term impact of cyberbullying.
Whereas offline bullying (or ``traditional'' bullying) relates to face-to-face interactions, cyberbullying typically takes place throughout a sequence of (intensive) user interactions in social media sites\cite{cheng2020session}, as illustrated in Fig. \ref{fig:intro}. 
Modeling user interaction is important to cyberbullying detection for several reasons. 
First, one essential aspect that differs cyberbullying from offline bullying is its \textit{persistence \& repetition}~\cite{ziems2020aggressive,dinakar2012common} characteristic. 
Decades of research in psychology~\cite{dooley2009cyberbullying} and social science \cite{smith2008cyberbullying} have shown that cyberbullying is carried out repeatedly against victims by one or more individuals. 
Studies of user interactions allow us to characterize the repetition by both content and temporal analysis. 
For example, recent findings~\cite{soni2018time,guptatemporal} suggest that (1) Most bullying comments occurred in the first hours after an initial post and the bullying comment counts in bullying sessions are significantly larger than those of non-bullying sessions; 
and (2) a relatively short time interval between consecutive cyberbullying comments is observed in bullying sessions.
Second, user interaction is a strong signal for detecting cyberbullying instances. 
Recent analysis of cyberbullying activity from an interdisciplinary research \cite{guptatemporal} manifests that the intensity of user activities within a bullying session is inherently different from that within a non-bullying session. 
Third, what makes cyberbullying detection particularly challenging compared to offline bullying is that it is not uncommon to see that users play multiple roles such as bully, victim, and witness in the same session~\cite{salmivalli1999participant,xu2012}. 
Therefore, modeling user interaction provides an alternative for the implicit monitor of users' roles over time. 

However, modeling user interaction in a social media session confronts multi-faceted challenges: 
(1) \textit{Sparsity}. 
Whereas a social media session is often involved with multiple interactions among numerous users, the interaction with a specific user can be very low. 
Further, most of the comments are directed towards the owner of the social media session, resulting in a ``hub'' where the owner degree is much larger than the degrees of users who post comments~\cite{ravazzi2017learning,huberman2009social}. 
Therefore, the user-user interaction graph constructed from reply relation can be sparse and less informative. 
(2) \textit{Repetition}. 
Characterizing the repetition of cyberbullying behavior is particularly challenging because we need to satisfy at least two constraints simultaneously: 
similar in content and close in time~\cite{soni2018time}. 
Therefore, the desired models should be able to jointly capture content similarities and temporal correlations. 
(3) \textit{User Characteristics}. Empirical findings within psychology identify that users' personality traits are important indicators of computer-mediated behaviors \cite{al2016cybercrime,kowalski2014bullying}. 
However, it is rather difficult to discover such user characteristics solely based on the content in a session due to the extremely limited comments from an individual user. 

To tackle these challenges, we propose \textbf{TGBully}, a \textbf{T}emporal-\textbf{G}raph-based cyber\textbf{bully}ing detection framework.
TGBully is composed of three modules, i.e., a \textit{semantic context modeling} module, a \textit{temporal graph interaction learning} module and a \textit{session classification} module.
In the first module, we contrast the standard content analysis of a lump of comments with a hierarchical construction of a sequence of comments, each of which comprises a sequence of words. 
To incorporate characteristics of users who post these comments, we further extract users' historical comments, from which we infer their language behavior. 
The \textit{temporal graph interaction learning} module implicitly constructs the user-user interaction graph by encoding the \textit{topic coherence} and \textit{temporal dynamics} of user comments in a graph attention network (GAT). 
In particular, we consider content similarities and time intervals between comments, and the user interactions are learned through the continuous update of the edge weights in the GAT. 
By doing so, TGBully models user interaction through the propagation of interactive signals in the constructed temporal graph.
The \textit{session classification} module aggregates the information representation learned from user interactions and performs the final session-level classification, i.e., determining whether a session is bullying or not.
To summarize, the main contributions of this work are:
% \begin{itemize}[leftmargin=*]
\begin{itemize}
     \item We make novel observations of dynamic user interactions reflected by the inherently related and continuously evolved comments in a social media session. We show the importance of modeling user interactions in improving the performance of cyberbullying detection. 
    \item We propose a principled and effective approach that models user interaction from two key aspects: temporal dynamics and topic coherence. This is achieved by a temporal graph constructed for each session and a GAT for learning the interactions in the graph. 
    \item We evaluate TGBully on both session-level detection tasks and comment-level case study.
    Empirical results on two benchmark datasets validate the effectiveness of our approach and shed light on its practical implications for policy-makers.
\end{itemize}
\section{Related Work}
Our work is closely related to cyberbullying detection and graph attention networks. Here, we briefly review the related literature.
\subsection{Cyberbullying Detection}
Previous work on cyberbullying detection typically builds their approaches upon natural language processing (NLP) methods~\cite{dinakar2011modeling,nand2016bullying,romsaiyud2017automated}.
Their goal is to identify aggressive and harassing mentions via text analysis.
Hosseinmardi et al.~\cite{hosseinmardi2014analyzing} compared Linguistic Inquiry and Word Count (LIWC) values of different categories of words, they found that words connected to negative feelings such as “depress”, “stressful”, and “suicide” are prominent identifiers of cyberbullying.
Dani et al.~\cite{dani2017sentiment} added sentiment consistency features to a unigram model with TF-IDF values as content features.

More recent work highlighted the importance of temporal information for cyberbullying detection~\cite{cheng2019hierarchical,soni2018time,cheng2020unsupervised}. 
For example, Soni and Sign~\cite{soni2018time} characterized each comment on Instagram by an exponential time factor.
They empirically evaluated a set of time-related features derived from the exponential factor and concluded several essential temporal traits of cyberbullying. Cheng et al.~\cite{cheng2019hierarchical,cheng2020unsupervised} unified cyberbullying detection and time interval prediction in a multi-task learning framework.
Built upon the hierarchical attention network, their approach aggregated contextualized comment representation for session classification meanwhile predicted time intervals from comment representations.
In addition to text, numerous methods incorporated social information into cyberbullying detection~\cite{huang2014cyber,chatzakou2017mean,ziems2020aggressive}. By correlating cyberbullying behavior with online sociability, they primarily measured the power imbalance between users by the  social network and interaction network.
Ziems et al.~\cite{ziems2020aggressive} utilized the neighborhood overlap, verified status, number of followers, and recent tweets to indicate users' visibility among peers.
Chatzakou et al.~\cite{chatzakou2017mean} extracted and compared numerous text, user, and network-based attributes. 
They found that compared with aggressors, bullies are less popular and participate in fewer online communities.
% However, social information is not always available due to privacy consideration. Therefore, this work seeks to develop cyberbullying detection models without social network information.

Prior research has put a large number of efforts into modeling user-generated contents but overlooked the interactions of users themselves within a session. Findings from various fields have evidenced the critical role user interactions play in cyberbullying detection. We thereby complement earlier work with the consideration of user interactions by jointly modeling the temporal dynamics and topic coherence in an established temporal graph.
\subsection{Graph Attention Network}
Recent years have witnessed the increasing popularity of graph neural networks (GNNs), which have become powerful tools to model real-world data in various domains~\cite{li2015gated,kipf2016semi,hamilton2017inductive}. Among the widely recognized GNNs, graph attention networks (GATs)~\cite{velivckovic2017graph} use self-attentional layers with automatically learned weights to attend over neighborhoods’ features. 
Due to their prominent capability of learning salient interaction, a surge of attempts have been made to apply GATs to various text mining tasks. For instance,
Hu et al.~\cite{linmei2019heterogeneous} proposed a Heterogeneous Graph Attention networks (HGAT) for semi-supervised short text classification. At its core, HGAT embeds heterogeneous information (topics, short texts, and entities) via a dual-level attention mechanism.
Huang et al.~\cite{huang2019syntax} designed a novel target-dependent graph attention network (TD-GAT) to explicitly utilize the dependency relation among words for aspect-level sentiment classification.

Our work is more related to another line of research exploiting GAT for fact verification~\cite{zhong2019reasoning,zhou2019gear,liu2020fine}. In both cases, characterizing interactions between facts (or users in our task) are central to the modeling process. 
For instance, Zhong et al.~\cite{zhong2019reasoning} constructed an XLNet-based word graph.
Then they formed argument-level representations using a graph convolution network to propagate over the word graph.
At last, GAT was incorporated to aggregate useful evidence for the verification task.
Similarly, Zhou et al.~\cite{zhou2019gear} also employed GAT to perform fact reasoning where each graph node was constructed by concatenating a claim and the corresponding evidence.
In these approaches, edge weights are automatically optimized during the propagation process.
However, these approaches are not applicable to our task due to the unique challenges of the session-based cyberbullying detection \cite{cheng2020session}. Therefore, to make the most of GAT, we propose to explicitly introduce the two important bully-indicative features -- temporal dynamics and topic coherence -- into the edge weight calculation.

\section{Problem Formulation}
We use the following information extracted from a social media session for cyberbullying detection:
\begin{itemize}
    \item \textbf{Comment: } It is a sequence of comments posted by different users in a session. 
    There are two types of comments in general: an initial comment posted by the owner and the subsequent comments.
    % For both types of comments, we represent words in comments as embeddings and analyze their contextual relation with neural networks.
    They comprise of the textual information of a session. 
    We represent words in a comment as embeddings and analyze their contextual relation using neural networks.
    \item \textbf{Time: }
    It is the time stamp of each comment. 
    Informed by previous work \cite{cheng2019hierarchical}, we represent time by calculating the time intervals (in minutes) between comments and the initial comment.
    \item \textbf{User: }
    It denotes the username of each posted comment. 
    % A typical scenario is a user may mention a previous username in his/her comments.
    Since a user may mention a previous username, we embed each username as a word and encode it into comment representation.
    % To capture this kind of interaction, we embed each username as a word and encode it into comment representation.
    We also add the username embedding before the first token of a comment to identify its poster.
    \item \textbf{History: }
    The historical comments of each user. 
    We collect historical comments from previous sessions and concatenate them into a unified paragraph.~\footnote{We omit their relative temporal order due to the lack of such information.}
\end{itemize}

Suppose that we have a corpus of sessions $\mathcal{C}$ and each session consists of $n$ comments, denoted as $\mathcal{S}=\{c_1, c_2, ..., c_n\}$, where $c_i$ is the $i^{th}$ comment in the session. 
Along with the sessions is the label set $Y$ indicating whether a session is a bullying ($y=1$) or non-bullying ($y=0$) session. 
A user $u\in\mathcal{U}$ posts comment $c_i=\{w_1^c, w_2^c, ..., w_p^c\}$ with $p$ words at time $t_i$. 
Besides, the user $u$ also has a set of historical comments, and we concatenate them into one paragraph with $q$ words as $h_u=\{w_1^h, w_2^h, ..., w_q^h\}, \forall h_u \in \mathcal{H}$.
We then define the session-based cyberbullying detection task as:

\noindent \textbf{Definition.} 
\textit{Given the corpus of sessions $\mathcal{C}$, the associated label set $Y$, the group of users $\mathcal{U}$, and the historical comments $\mathcal{H}$, the goal of this work is to classify a social media session into the bullying/non-bullying category using the multi-modal information extracted from Comment, Time, User and History. }

% Instead of simply concatenating information from different modalities, we explicitly capture bullying specific characteristics from the synergy of multi-modal attributes.
% We then formulate our task as below:

\section{TGBully: The Proposed Approach}
\begin{figure*}[t]
	\centering
	\resizebox{0.88\textwidth}{!}{\includegraphics{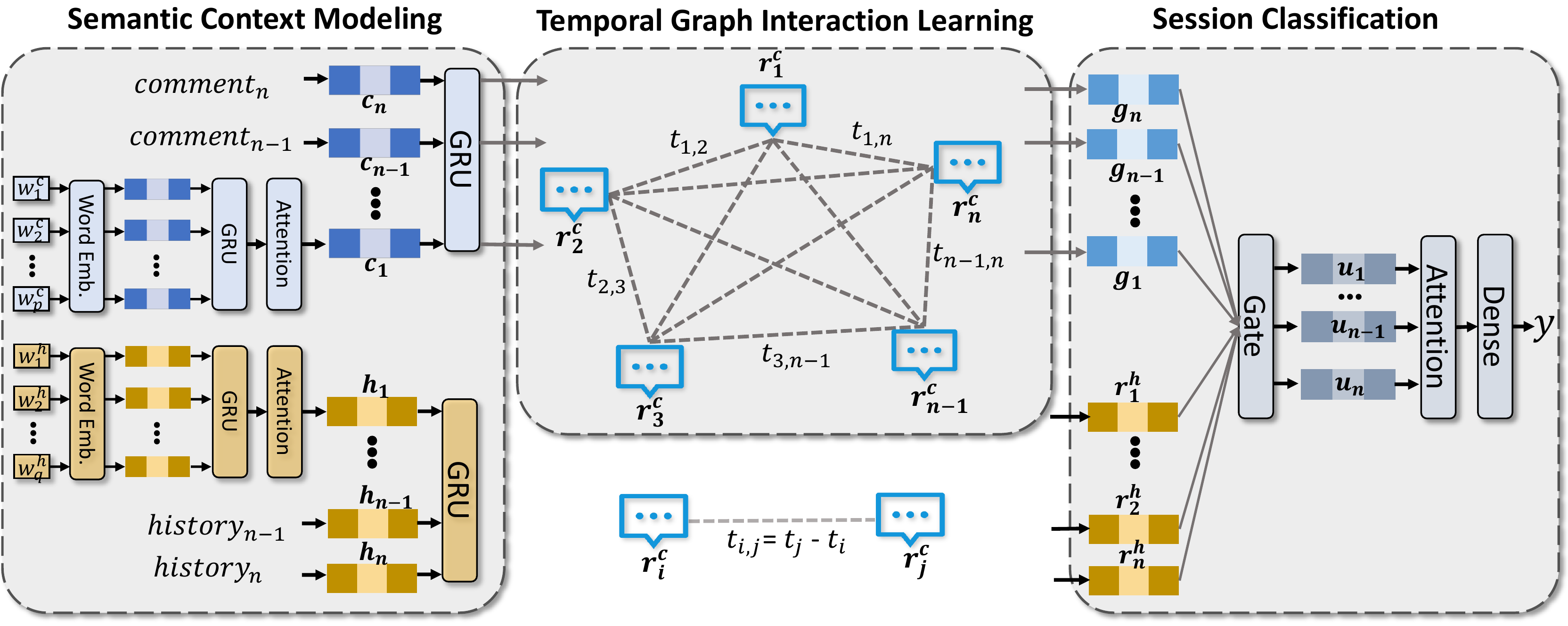}}
	\caption{Description of the overall framework. 
	TGBully consists of three modules:
	The \textit{semantic context modeling} module (left) hierarchically constructs comment sequences as $\mathbf{c}$ (blue) and encodes users' language behavior via the representation of historical comments  $\mathbf{h}$ (yellow).
	The \textit{temporal graph interaction learning} module (middle) implicitly encodes the \textit{topic coherence} and \textit{temporal dynamics} in the temporal interaction graph where nodes represent user comments and edges represent time intervals between two comments.
    The \textit{session classification} module (right) first aggregates the information learned from user interactions into user representation $\mathbf{u}$ (grey).
    It then performs the session-level classification by aggregating all user representations within the session into the session representation.
}
	\label{fig:model}
\end{figure*}

% In this section, we introduce our proposed cyberbullying detection method \textbf{GBully} in detail.
% The overall framework in Fig.~\ref{fig:model} shows that GBully consists of three modules, i.e., (1) a \textit{semantic context modeling} module to represent each comment from its textual content and user's historical comments, (2) a \textit{temporal graph interaction learning} module to encode comment interaction with a graph attention network.
% The GAT jointly captures \textit{topic coherence} and \textit{temporal dynamics} through its edge weights.
% (3) a \textit{session classification} module to attentively aggregate comment and history representations, and classify sessions based on them.
% Details of each module are introduced in the following sections.
The goal of this work is to study user interaction in a session to improve the session-level cyberbullying detection. We herein propose \textbf{TGBully}, a principled framework consisting of three modules: (1) a \textit{semantic context modeling} module that encodes each comment by considering both its textual content and user's language behavior reflected from her/his historical comments, (2) a \textit{temporal graph interaction learning} module that constructs a temporal graph and models the dynamic user interaction with a bully-featured GAT. The proposed GAT jointly captures \textit{topic coherence} and \textit{temporal dynamics} in the modeling process; and (3) a \textit{session classification} module that attentively aggregates information from user interaction into a session representation, based on which it then classifies the session into a bullying/non-bullying session.
The overall framework of TGBully is illustrated in Fig.~\ref{fig:model}.

\subsection{Semantic Context Modeling}
Towards the semantic understanding of texts in a session, the \textit{semantic context modeling} module utilizes a hierarchical pipeline~\cite{yang2016hierarchical,cheng2019hierarchical} to form two types of contextualized representations.
One is comment representation, which uncovers user intent and emotion through their utterances in the current session.
The other is history representation, which encodes user characteristics, e.g., personalities and language patterns reflected from their historical comments.
\subsubsection{Comment Encoding}
The comment encoding mainly consists of four layers. The first layer is the word embedding, which transforms a $p$-word comment $c=\{w_1^c, w_2^c, ..., w_p^c\}$ to a sequence of high dimensional vectors $\{\mathbf{w}_1^c, \mathbf{w}_2^c, ..., \mathbf{w}_p^c\}$ via an embedding matrix.
% It converts a sequence of words into a sequence of low-dimensional dense vectors which contain semantic information of these words. 
% Denote a comment $c$ containing $p$ words as $c=\{w_1^c, w_2^c, ..., w_p^c\}$, it is transformed to a sequence of high dimensional vectors $\{\mathbf{w}_1^c, \mathbf{w}_2^c, ..., \mathbf{w}_p^c\}$ via a embedding matrix.
% The word embedding matrix $\mathbf{E}$ is initialized using pre-trained word embeddings, and fine-tuned during model training.
The second is a word-level gated recurrent units (GRU)~\cite{cho2014learning} layer.
It aims to encode long-term contextual dependency among words.
% Since GRU reduces parameters by removing the output gate, it has been proven to showcase better performances than typical LSTM~\cite{hochreiter1997long} on some small datasets~\cite{gruber2020gru}.
It has been shown that GRU achieves better performance than typical LSTM~\cite{hochreiter1997long} on small datasets~\cite{gruber2020gru}, therefore, we use a stack of bidirectional GRUs to model word sequences from both forward and backward directions. 
% Considering cyberbullying datasets are relatively small in scale, we choose a stack of bidirectional GRUs to model word sequences from both forward and backward directions. 
The contextualized representation $\mathbf{r}_i^w$ of the $i^{th}$ word $\mathbf{w}_i^c$ is calculated as:
\begin{equation}
    \mathbf{r}_i^w=[\overrightarrow{GRU}(\mathbf{w}_i^c), \overleftarrow{GRU}(\mathbf{w}_i^c)].
\end{equation}
However, the words in a comment are not equally significant to cyberbullying. 
To automatically select informative words from a comment, we herein employ word-level attention as the third layer. 
% Online social media comments may include much user generated noise, thus different words in the sentences usually have different importance for bully detection.
% For instance, in the sentence ``youuu silly little boy , u know what hhh'', the word ``silly'' is more bully-indicative then other words.
% To automatically select useful words of a comment, our model use a word-level attention network.
For the $j^{th}$ comment, the aggregated comment representation $\mathbf{c}_j$ is learned as:
% \begin{equation}
%     a_i^w=\tanh(\mathbf{v}_w^T \times \mathbf{r}_i^w+b_w),
% \end{equation}

\begin{equation}
    \alpha_i^w=\frac{\exp(\tanh(\mathbf{v}_w^T \times \mathbf{r}_i^w+b_w))}{\Sigma_{k=1}^{p} \exp(\tanh(\mathbf{v}_w^T \times \mathbf{r}_k^w+b_w))},
\end{equation}

\begin{equation}
    \mathbf{c}_j=\displaystyle \sum_{i=1}^p \alpha_i^w \mathbf{r}_i^w,
\end{equation}
where $\mathbf{v}_w$ and $b_w$ are the parameters of the attention network.
$\alpha_i^w$ is the attention weight of the $i^{th}$ word.
The final layer of this sub-module consists of comment-level bidirectional GRUs that model comments in a temporal order.
Then the bi-GRU output representation of the $j^{th}$ comment is formulated as:
\begin{equation}
    \mathbf{r}_j^c=[\overrightarrow{GRU}(\mathbf{c}_j), \overleftarrow{GRU}(\mathbf{c}_j)].
\end{equation}

\subsubsection{History Encoding} As the number of comments posted by a user is rather limited, the goal of history encoding is to better capture users' language behavior and personality traits mirrored by their historical comments.
Survey studies find that bullying behavior has strong connections with both hostility and neuroticism~\cite{aricak2009psychiatric,corcoran2012cyberbullying}. Neurotic users are more emotional and sometimes even more irrational, thereby, they are more likely to engage in cyberbullying activities~\cite{al2016cybercrime}. 
Further, it has been advocated that the use of historical information in social media is effective for personality prediction~\cite{golbeck2011predicting}.  
We first associate the learned user characteristics with corresponding comments in the current session. 
We then use the same semantic context modeling approach described above for user history modeling. In particular,
given $n$ users' history in a session as inputs, where each history comment comprises $q$ words, the output contextualized history sequence is $\{\mathbf{r}_1^h, \mathbf{r}_2^h, ..., \mathbf{r}_n^h\}$.

\subsection{Interaction Learning in Temporal Graph}

The \textit{Temporal graph interaction learning} module seeks to learn user interaction and capture the repetitive characteristics of cyberbullying behavior.
To accomplish this goal, the module first constructs a temporal interaction graph for each social media session.
Since GAT can selectively aggregate principle neighbor information via its attentional layers, we use it to implicitly encodes the \textit{topic coherence} and \textit{temporal dynamics} between different users in the graph.

We begin by illustrating how the \textit{temporal graph interaction learning} module can effectively address the two major challenges for modeling user interactions in a session.
To recall, as most comments are directed towards the owner of the session, the first challenge is that social graph based on reply and friendship relation can be sparse; the second challenge relates to the unique characteristic of cyberbullying -- repetition: content similarity and time proximity. We argue that these challenges can be effectively tackled by jointly modeling the \textit{topic coherence} and the \textit{temporal dynamics} in a session. The two aspects improve our understandings of the discussed topics and temporal correlation across time.

We first construct the temporal graph where the nodes are user comments and the edges are time intervals between comments. Each comment in our graph is inherently connected, consequently, they will not be bounded by sparse and uneven reply relationships between their users. We then propose to encode the topic coherence and temporal dynamics in the temporal graph jointly in a unified framework. 
The affinity level of two neighbor user comments, namely, their content similarity and time proximity, is automatically determined and propagated through the edges of the temporal graph. At its core, the \textit{temporal graph interaction learning} module leverages GATs to aggregate neighbor information (i.e., comments) according to their affinity levels.

\subsubsection{Temporal Graph} The user interaction is implicitly learned by capturing their comment interaction in the temporal graph.
Specifically, we first transform the session into a fully connected graph, where we establish connectivity between two arbitrary comments.
% Specifically, we exploit a fully connected Graph Attention Network (GAT)~\cite{velivckovic2018graph} for each session to capture the interconnection between each pair of comments.
% With our proposed GAT, the vanishing gradient problem is naturally solved by establishing connectivity between two arbitrary comments.
In the temporal graph, comments are viewed as graph nodes and time intervals between them are viewed as edges. 
Informed by recent advances in GNNs~\cite{huang2019syntax,linmei2019heterogeneous}, we incorporate a graph attention neural network~\cite{velivckovic2018graph} to automatically aggregate information from neighbor nodes to the central node.
Different significance levels of neighbor nodes are weighted by their edge weights.
The GAT network encodes the temporal-level and topic-level relatedness between the central node and neighbors by introducing them into the calculation of edge weights.
Then it propagates interactive signals to the central nodes by aggregating neighbors nodes according to their edge weights. Additionally, we consider self-propagation to reserve the original content information of each node.
Since our temporal graph is already fully-connected, instead of stacking multiple graph layers, we only perform one-hop graph aggregation\footnote{Empirical experimental results also prioritize the one-hop graph aggregation.}.
For the $j^{th}$ comment in the temporal graph, we denote its node vector as $\mathbf{g}_j$, which is calculated as:
\begin{equation}
    \mathbf{g}_j= \sum_{\mathbf{r}_k^c \in \{N_j \cup \mathbf{r}_j^c\}} \pi(\mathbf{r}_k^c,\mathbf{r}_j^c,t_{k,j})\mathbf{W}_c\mathbf{r}_k^c,
\end{equation}
where $N_j$ is the set of $j^{th}$ comment's neighbor nodes, $\mathbf{W}_c$ is the transformation matrix of the graph network.
$t_{k,j}=t_j-t_k$ is the time interval between two comments.
$\pi(\mathbf{r}_k^c,\mathbf{r}_j^c,t_{k,j})$ denotes the weight of edge connecting node $\mathbf{r}_k^c$ and $\mathbf{r}_j^c$.
It is a decaying factor that controls the relative contribution each neighbor makes to the central node.
We detail the implementation of $\pi(\mathbf{r}_k^c,\mathbf{r}_j^c,t_{k,j})$ later.
\subsubsection{Topic coherence}
Semantic relatedness between comments in a session can reveal users' reactions and attitudes towards the commenting behavior of other users.
For instance, a group of users may repeat the offensive terms used in a previous post to humiliate the victim.
Such language repetition reflects and reinforces their aggressive motivations and potential power imbalance between bullies and victims. 
Crucially, capturing such topic-level repetition is useful for cyberbullying detection.

Our proposed approach is based on an important observation:
In the embedding space, two comments usually locate in close proximity if they share similar content or they are coherent in the topical discussion. To measure such proximity, we compute the embedding similarity $\pi_{time}$ of comments by taking dot products of their transformed node vectors in the graph. Formally,
\begin{equation}
   \pi_{topic}(\mathbf{r}_k^c,\mathbf{r}_j^c)=(\mathbf{r}_k^c\mathbf{W}_o)^T\mathbf{r}_j^c,
\end{equation}
where $\mathbf{r}_k^c$ and $\mathbf{r}_j^c$ are two nodes representing two arbitrary user comments. $\mathbf{W}_o$ is the graph transformation matrix.
\subsubsection{Temporal dynamics}
In addition to the topic coherence, temporal proximity is also an important feature for repetition. 
Instead of being immediate one-off responses, previous research discovered that cyberbullying sessions receive a more steady stream of comments that are consecutive in time~\cite{soni2018time}.
It is also common to observe bursts of comments in which several bully users gang up on a victim in a bullying session~\cite{al2013cyber}.
Both the steady stream of replies and a sudden burst of comments can be recognized by capturing intervals between each chronological pair of comments.
Thus, we further integrate the time-lapse factor into the constructed temporal graph.
Specifically, we calculate the time interval of comments and project it into the graph embedding space as below:
\begin{equation}
   \pi_{time}(t_{k,j})=W_t t_{k,j}=W_t (t_j-t_k),
\end{equation}
where $W_t$ is the time transformation coefficient and $t_{k,j}$ denotes the time interval.
\subsubsection{Joint modeling}
The ultimate goal of this work is to improve cyberbullying detection with user interaction modeling. Here, we detail how to learn the topic coherence and temporal dynamics of user interaction in a unified framework such that the cyberbullying detection performance can be optimized. 
% identify the repetitive characteristic of cyberbully through the combination of attributes from multiple modalities. 
% Existing efforts achieved this through various techniques, e.g., feature engineering~\cite{chatzakou2017mean}, multi-task learning~\cite{cheng2019hierarchical}, and network embedding~\cite{cheng2019xbully}.
% However, limited exploration has been done to encode different attributes into comment-comment relation modeling.
% Different from the above techniques, we integrates \textit{topic coherence} and \textit{temporal dynamics} into the modeling process of \textit{user interaction} so as to gain more comprehensive understanding of bullying behavior.
One empirical observation is that the affinity level of  two comments should be enhanced when they:
(1) discuss consistent and coherent topics; and 
(2) appear closely in the time scope. Therefore, we integrate the topical factor $\pi_{topic}$ and the temporal factor $\pi_{time}$ into the learning of the edge weight $\pi(\mathbf{r}_k^c,\mathbf{r}_j^c,t_{k,j})$:
\begin{equation}
%   \pi(\mathbf{r}_k^c,\mathbf{r}_j^c,t_{k,j})=\tanh((\mathbf{r}_k^c\mathbf{W}_o)^T\mathbf{r}_j^c+W_t t_{k,j}),
   \pi(\mathbf{r}_k^c,\mathbf{r}_j^c,t_{k,j})=\tanh(\pi_{topic}(\mathbf{r}_k^c,\mathbf{r}_j^c)+\pi_{time}(t_{k,j})).
\end{equation}
\noindent Note that for simplicity, we only utilize dot products of two nodes and add the results to the transformed time intervals.
Further explorations of the edge weight calculation, such as alternative forms of attention mechanism and different ways of combining the two components, are left as future work. By modeling user interactions via the bully-featured GAT -- GAT that encodes topic coherence and temporal dynamic jointly, we mimic the information flow in the social network to capture the evolution of the discussed topics and the repetitive characteristics of cyberbullying.

\subsection{Session Classification}
The \textit{session classification} module aims to classify a social media session into a bullying or non-bullying session. In particular, it first aggregates user's history representation $\mathbf{r^h}$ and graph representation $\mathbf{g}$ into a unified user representation $\mathbf{u}$.
Then it incorporates a user-level attention network to select informative users to form the final session representation $\mathbf{s}$, which is used for the final classification task.

While examining users' historical comments, an important notion is that some users show similar language behavior over time.
For example, a user who constantly bullies others comments with offensive remarks whenever s/he is involved in an online discussion.
By contrast, a user who occasionally bullies others may make offensive comments only once.
Under such circumstance, her/his historical comments may not be informative enough to recognize the user's bullying tendency in the current session.
Considering the relative importance of historical information for each user, we here use a gating mechanism to automatically weigh the relative importance of comments in the current session and user history.
For the $j^{th}$ comment in session $s$, the merged user representation $\mathbf{u}_j$ is formulated as below:
\begin{equation}
   \mathbf{u}_j=\beta_j \odot \mathbf{r}_j^h+(1-\beta_j) \odot \mathbf{g}_j,
\end{equation}
\begin{equation}
   \beta_j=\sigma(\mathbf{W}_h \mathbf{r}_j^h+\mathbf{W}_g \mathbf{g}_j+b_c),
\end{equation}
where $\mathbf{W}_h$, $\mathbf{W}_g$ and $b_c$ are trainable parameters, $\sigma$ is the sigmoid activation function, and $\beta_j$ is the controlling weight.
Since different user comments are not equally relevant to cyberbullying detection, we aggregate them using a user-level attention mechanism.
The final session representation $\mathbf{s}$ is learned as:
% \begin{equation}
%     a_j^s=\tanh(\mathbf{v}_s^T \times \mathbf{u}_j+b_s),
% \end{equation}

\begin{equation}
    \alpha_j^s=\frac{\exp(\tanh(\mathbf{v}_s^T \times \mathbf{u}_j+b_s))}{\Sigma_{k=1}^{n} \exp(\tanh(\mathbf{v}_s^T \times \mathbf{u}_k+b_s))},
\end{equation}

\begin{equation}
    \mathbf{s}=\displaystyle \sum_{j=1}^n \alpha_j^s \mathbf{u}_j,
\end{equation}
where $\mathbf{v}_s$ and $b_s$ are parameters of the attention network, and $\alpha_j^s$ is the attention weight of the $j^{th}$ user. For session-level cyberbullying detection, we feed $\mathbf{s}$ into a single-layer dense network and predict its label $y$, where $y=1$ indicates a bullying case.

\section{Experiments}
In this section, we begin by an introduction of the benchmark datasets for session-based cyberbullying detection. We then detail the experimental setup including the baselines, evaluation metrics , and implementations. The focus of this section lies in answering the following questions:
\begin{itemize}
    \item \textbf{RQ1:} How does TGBully fare against existing cyberbullying detection models w.r.t. predictive performance?
    \item \textbf{RQ2:} How does each module of TGBully contribute to the overall model performance?
    \item  \textbf{RQ3:} As real-world cyberbullying datasets are often imbalanced, how does the data imputation techniques (e.g., oversampling) influence the session-level cyberbullying detection?
    \item \textbf{RQ4:} Apart from the session-level prediction, can TGBully effectively recognize cyberbullying instances at the comment-level?
\end{itemize}
\subsection{Data}
\label{annotation}
\begin{table}
    \centering
    \caption{Statistics of the Instagram and Vine dataset.}
\resizebox{0.48\textwidth}{!}{
    \begin{tabular}[t]{|c||cccc|}
         \hline
         Datasets & \textbf{\# Sessions} & \textbf{\# Bully} & \textbf{\# Non-bully} & \textbf{\# Comments}\\
         \hline
         \textit{Instagram} & 2,218 & 678 & 1,540 & 155,260\\
        %  \hline
         \textit{Vine} & 970 & 304 & 666 & 78,250\\
         \hline
     \end{tabular}
}
    \label{tab:dataset}
\end{table}
We conduct experiments on two benchmark datasets collected from Instagram~\cite{hosseinmardi2015analyzing} and Vine~\cite{rafiq2015careful}.
Both datasets were crawled using a snowball sampling method and manually annotated via the crowd-sourcing platform CrowdFlower.
Definitions of cyberbullying were provided for five annotators at the beginning of the labeling process.
A session is considered a cyberbullying session if no less than three people labeled it as bully.  
Sessions were removed from the dataset if they contain less than 15 comments.
Statistics of the two datasets are illustrated in Table~\ref{tab:dataset}.

\noindent\textbf{Instagram:} 
Instagram\footnote{https://www.instagram.com/} is a media-based social network where users post, like, and comment on images. 
It has been reported as one of the top five networks with the highest percentage of users reporting experience of cyberbullying~\cite{Ditch}.
For each media-session, this dataset provides the labeled image content, the textual content of comments and the arrival time of comments. 
In addition, the user of each comment is identified along with social network properties (user IDs of their likes, followers, and followings) and historical information (comments and likes in earlier sessions).
In total, this dataset contains 2,218 media sessions with an average number of 72 comments in each session. 
For each comment, the average number of words is 12.

\noindent\textbf{Vine:} Vine\footnote{https://vine.co/} is a mobile application that allows users to upload, like, and comment on six-second looping videos.
For each video session, along with the video content, this dataset also records content, time, and associated user IDs of each comment.
However, different from the Instagram dataset, this dataset doesn't provide users' historical comments. 
Instead, it provides user profile information such as location, profile description, and social network properties (user IDs of their followers and followings).
In implementation, we substitute user historical comments in the history encoding module with their profile description.
On average, there are 81 comments per session and 8 words per comment.

\begin{table*}[t]
\centering
\caption{Evaluation results of different methods on the Instagram dataset.}
\resizebox{0.85\textwidth}{!}{
\begin{tabular}{|c||ccc|ccc|}
\hline
             &\multicolumn{3}{c|}{\textbf{50\%}} & \multicolumn{3}{c|}{\textbf{100\%}} \\ \hline
            & \small{\textbf{Recall}}& 
             \textbf{F1} & 
             \textbf{AUC} & 
             \small{\textbf{Recall}} & 
             {\textbf{F1}} &
             \textbf{AUC}
             \\ \hline

LR        &52.74$\pm$4.48&64.05$\pm$2.93&73.89$\pm$1.94&59.06$\pm$3.79&69.12$\pm$2.86&76.97$\pm$1.75\\
SVM       &55.93$\pm$4.80&66.31$\pm$3.36&75.28$\pm$2.20&59.32$\pm$3.85&69.03$\pm$2.94&76.94$\pm$1.77\\
XGBoost   &69.05$\pm$5.71&72.25$\pm$1.59&79.63$\pm$1.16&71.55$\pm$3.64&75.25$\pm$2.09&81.68$\pm$1.57\\
\hline
CNN        &67.41$\pm$4.53&66.67$\pm$0.95&83.39$\pm$1.15&70.51$\pm$5.84&67.12$\pm$1.53&84.00$\pm$1.37\\
LSTM        &64.18$\pm$2.24&66.63$\pm$1.37&82.22$\pm$2.45&70.01$\pm$8.37&67.42$\pm$1.60&83.87$\pm$1.59\\
SelfAtt        &66.73$\pm$6.24&72.17$\pm$5.40&90.30$\pm$1.92&74.05$\pm$4.88&74.45$\pm$3.00&89.32$\pm$1.37\\
HAN        &71.69$\pm$4.24&75.01$\pm$1.74&90.07$\pm$1.40&78.21$\pm$7.02&76.99$\pm$1.99&91.16$\pm$1.02\\
\hline
SICD &62.53$\pm$4.53&69.71$\pm$2.38&79.92$\pm$1.27&76.87$\pm$3.79&70.55$\pm$1.29&86.24$\pm$1.16\\
Soni \& Sign &65.67$\pm$5.32&70.68$\pm$1.87&84.19$\pm$0.82&72.06$\pm$4.01&72.59$\pm$1.43&88.06$\pm$1.24\\
HANCD  &\textbf{78.39$\pm$3.69}&76.40$\pm$2.17&91.63$\pm$2.13&76.91$\pm$4.56&77.55$\pm$2.14&91.32$\pm$1.29\\
\textbf{TGBully}&77.64$\pm$4.70&\textbf{78.12$\pm$2.04}&\textbf{91.81$\pm$0.98}&\textbf{82.57$\pm$3.10}&\textbf{80.97$\pm$2.03}&\textbf{92.91$\pm$1.30}\\
\hline

\end{tabular}
}
\label{ins_result}
\end{table*}

\begin{table*}[t]
\centering
\caption{Evaluation results of different methods on the Vine dataset.}
\resizebox{0.85\textwidth}{!}{
\begin{tabular}{|c||ccc|ccc|}
\hline
             & \multicolumn{3}{c|}{\textbf{50\%}} & \multicolumn{3}{c|}{\textbf{100\%}} \\ \hline
             & \small{\textbf{Recall}}  &
             \textbf{F1} & 
             \textbf{AUC} & 
             \small{\textbf{Recall}}     &
             {\textbf{F1}} &
             \textbf{AUC} 
             \\ \hline

LR        &47.42$\pm$5.32&55.92$\pm$3.21&63.72$\pm$2.25&50.00$\pm$4.06&59.65$\pm$3.31&65.85$\pm$2.30\\
SVM        &47.69$\pm$4.98&57.22$\pm$3.17&64.31$\pm$2.42&48.48$\pm$3.28&60.38$\pm$2.76&66.81$\pm$1.78\\
XGBoost        &63.11$\pm$4.95&61.32$\pm$2.89&69.82$\pm$1.88 &64.54$\pm$3.77&64.00$\pm$3.12&70.75$\pm$1.42\\
\hline
CNN        &60.50$\pm$6.90&53.97$\pm$2.43&67.07$\pm$0.70&61.61$\pm$2.27&56.56$\pm$1.40&72.65$\pm$2.01\\
LSTM        &60.12$\pm$8.23&57.76$\pm$1.78&73.64$\pm$4.17&62.76$\pm$9.77&58.91$\pm$2.61&73.04$\pm$5.05\\
SelfAtt        &53.00$\pm$7.29&59.46$\pm$4.33&81.77$\pm$1.60&59.83$\pm$6.56&64.49$\pm$3.18&81.39$\pm$1.74\\
HAN        &71.11$\pm$6.28&66.54$\pm$2.11&82.13$\pm$1.77&70.04$\pm$2.93&68.10$\pm$2.26&82.32$\pm$1.18\\
\hline
SICD &61.72$\pm$5.24&58.43$\pm$5.75&76.74$\pm$1.64&66.82$\pm$3.92&62.76$\pm$2.38&78.56$\pm$2.52\\
Soni \& Sign &64.50$\pm$5.98&60.47$\pm$2.43&80.27$\pm$1.35&64.48$\pm$5.24&63.49$\pm$1.97&80.33$\pm$3.27\\
HANCD  &68.39$\pm$3.94&67.10$\pm$3.08&\textbf{83.62$\pm$2.08}&\textbf{71.05$\pm$4.66}&68.01$\pm$2.86&82.76$\pm$3.13\\
\textbf{TGBully}&\textbf{75.17$\pm$6.52}&\textbf{68.64$\pm$3.78}&83.54$\pm$3.35&70.18$\pm$6.65&\textbf{69.35$\pm$2.04}&\textbf{83.05$\pm$2.67}\\
\hline
\end{tabular}
}
\label{vine_result}

\end{table*}

\subsection{Experimental Setup}
For each dataset, we randomly sample 80\% sessions for training, 10\% for validation, and 10\% for testing.
We assume each session is independent and do not consider cross-session correlation.
Session length is set as 140 and sentence length is 30.
We use the pretrained 400-dim Twitter embedding~\cite{godin} as the textual input.
The sentence-level Bi-GRU has 2$\times$32 hidden states and the session-level Bi-GRU has 2$\times$64.
The dropout rate is 0.2.
We repeat each experiment 5 times and report their average performances and standard deviations.

\noindent \textbf{Evaluation Metrics.}
We use three widely recognized evaluation metrics in cyberbullying detection tasks, i.e., recall, micro F1, and AUC.
We pay special attention to recall scores as the frequency of bullying sessions in the real-world is typically much lower compared to that of non-bullying sessions.
Further, we report the micro F1 score for the interest of cyberbullying instances.

\noindent \textbf{Baselines.}
We compare TGBully with three categories of baselines: \textit{feature-based models} (LR, SVM and XGBoost), \textit{neural text classification models} (CNN, LSTM, SelfAtt and HAN), and \textit{cyberbullying detection models} (SICD, Soni \& Sign and HANCD). Specifically,
\begin{itemize}
    \item \textbf{LR.} Logistic Regression is a commonly used binary classifier. 
    It estimates and optimizes parameters of a logistic model.
    \item \textbf{SVM}~\cite{cortes1995support}. 
    Support Vector Machine (SVM) is a non-probabilistic binary classifier.
    It maximizes the functional margin between positive and negative classes.
    \item \textbf{XGBoost}~\cite{chen2016xgboost}.
    It is a highly scalable and efficient tree boosting system.
    The decision tree-based algorithm uses several regularization and parallelization techniques to improve overall performances.
    \item \textbf{CNN}~\cite{kim2014convolutional}.
    The original implementation uses convolutional neural network (CNN) to classify single sentences.
    We apply it at comment level and average comment representations to form session representations.
    \item \textbf{LSTM}~\cite{hochreiter1997long}.
    It designs long short-term memory units for context modeling.
    Similarly, we use it to model comments and classify sessions with averaged comment representations.
    \item \textbf{SelfAtt}~\cite{vaswani2017attention}.
    It incorporates multi-head self-attention mechanism to model context dependency in sentences.
    In addition to word-level attention, we apply a similar comment-level attention to model relation between different comments.
    \item \textbf{HAN}~\cite{yang2016hierarchical}.
    A hierarchical bi-directional GRU framework is used to model sessions at both word level and comment level.
    \item \textbf{SICD}~\cite{dani2017sentiment}.
    A sentiment enhanced feature-based cyberbullying model.
    User-post relationship and sentiment similarity between posts are used to guide the prediction.
    \item \textbf{Soni \& Sign}~\cite{soni2018time}.
    A cyberbullying model that exploits temporal information and activity levels in sessions. 
    Textual, social and temporal features are simultaneously utilized for classification.
    \item \textbf{HANCD}~\cite{cheng2019hierarchical}.
    A multi-task cyberbullying model based on HAN.
    It adds a time interval prediction subtask to exploit temporal dynamics within sessions.
\end{itemize}

For LR, SVM, and XGBoost, we use word-level tf-idf of user comments as textual features and number of likes and follows as social features.
We use the same word embedding for TGBully and all \textit{neural text classification models}.
Hyperparameters are tuned to optimize performance.
We set the parameters as reported in the original papers for all \textit{cyberbullying detection models}.
One of the challenges in cyberbullying detection is the data scarcity problem.
Therefore, to investigate how much these models rely on the labeled data, we also construct another training set by randomly sampling half of the original training data. We report results for both cases with half and full training data. 
\subsection{Performance Evaluation}
To answer the question in \textbf{RQ1}, we compare the performances of all baseline methods with TGBully's on the Instagram and Vine datasets. 
The results are reported in Table~\ref{ins_result} and~\ref{vine_result}. 

Comparing TGBully with other cyberbullying baselines that also consider temporal analysis, we observe that: 
(1) TGBully consistently achieves the best performances in most cases with different datasets and training ratios. This is in part because our approach effectively models user interactions in a graph attention neural network with a focus on temporal dynamics and topic coherence of a session.
The joint modeling of the two bully-indicative aspects further leads to better recognition of the repetitive characteristics of cyberbullying behavior.
The improvement also comes from the modeling of user personality traits and language behavior from users' historical comments while learning user interaction.
(2)
Large variations are observed in the results of all methods, especially when there is less training data.
This reveals the necessity of using larger datasets in order to further improve the performance of cyberbullying detection. 
This result is also evidenced by recent findings in~\cite{cheng2020session}, which states that 
TGBully improves the performance by a smaller margin over baselines with the Vine dataset as Vine has fewer samples and we do not have access to the historical comments of its users.

By comparing different methods' performances, we also conclude that: 
(1) Neural models achieve substantially better performances than feature-based methods, especially w.r.t. Recall and AUC.
Rather than representing session information with raw features, neural networks incorporate pretrained word embeddings that have rich semantic meanings and are fine-tuned during the training process. Another important reason is that neural models can better capture context dependency between words, leading to high-quality session representations, even with highly imbalanced datasets.
(2) Another observation is that explicitly modeling the hierarchical structure can further improve the performance.
Hierarchical neural models (SelfAtt, HAN, HANCD, and TGBully) outperform standard neural models (CNN and LSTM) partly because they better capture the interaction between different comments and consider their relative importance in a session.
(3) Introducing additional temporal and sentiment information enhances both \textit{feature-based models} and \textit{text classification models}. 
For instance, SICD and Soni \& Sign introduce sentiment and temporal modeling respectively, and both methods outperform LR and SVM methods by a large margin.
Similarly, HANCD and TGBully improve over HAN model by modeling the temporal dynamics in a session.

\begin{figure}[t]
	\centering
	\resizebox{0.37\textwidth}{!}{\includegraphics{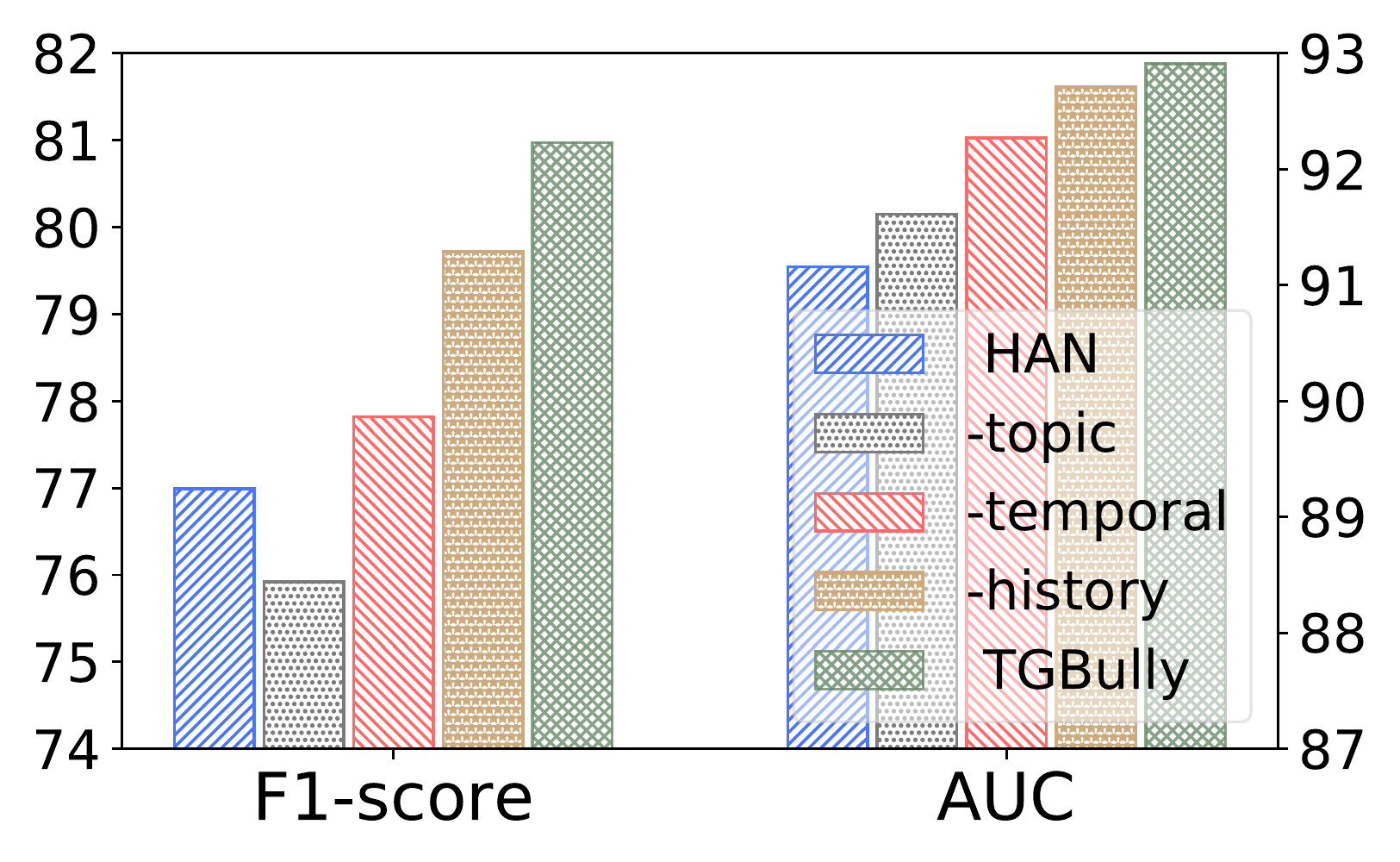}}
	\caption{TGBully and its variants on the Instagram dataset.}
	\label{fig:ablation}
\end{figure}

\begin{figure*}[t]
\centering
\subfigure[A bullying example.]{
\begin{minipage}[t]{0.8\linewidth}
\centering
\includegraphics[scale=0.4]{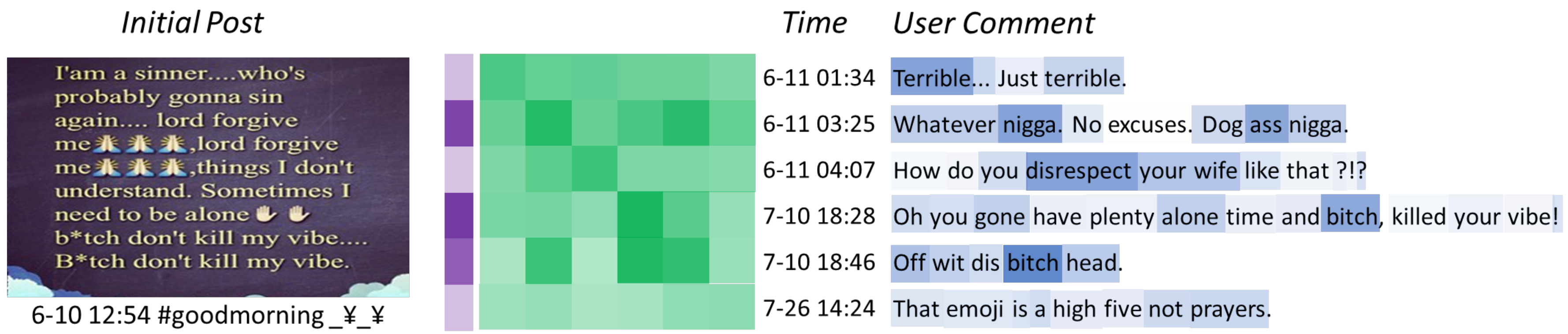}
\label{fig:bully}
\end{minipage}
}
\subfigure[A non-bullying example.]{
\begin{minipage}[t]{0.8\linewidth}
\centering
\includegraphics[scale=0.4]{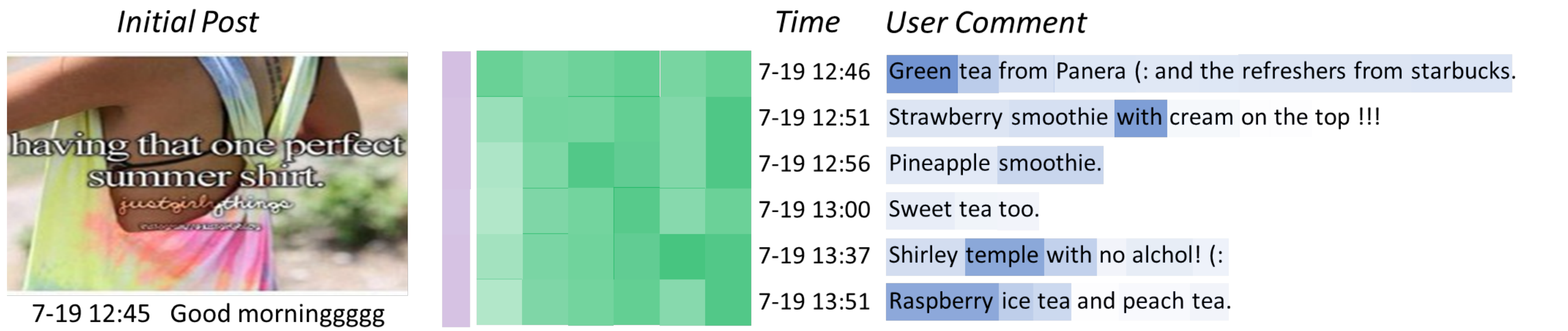}
\label{fig:nonbully}
\end{minipage}
}
\centering
\caption{Visualization of various types of weights within a session. 
Shades of purple and blue denote user-level and word-level attention weights respectively.
Green mesh grids denote graph weights between users (i.e., their comments), where grid at the $i^{th}$ row and $j^{th}$ column represents weight $w_{ij}$.
Darker color represents higher weights.
}
\label{fig:casestu}
\end{figure*}

% \begin{figure*}[t]
% 	\centering
% 	\resizebox{0.8\textwidth}{!}{\includegraphics{fig/case_study.pdf}}
% 	\caption{Visualization of various types of weights in a cyberbullying session.}
% 	\label{fig:case1}
% \end{figure*}

% \begin{figure*}[t]
% 	\centering
% 	\resizebox{0.8\textwidth}{!}{\includegraphics{fig/case_study_2.pdf}}
% 	\caption{Visualization of various types of weights in a non-bullying session.}
% 	\label{fig:case2}
% \end{figure*}

\subsection{Ablation Study}
This set of experiments seeks to examine the impact of each module of TGBully on the overall performance (\textbf{RQ2}).
Here, we use the Instagram dataset as a working example because similar findings can be observed with Vine. Results are reported in Fig.~\ref{fig:ablation}.
We use the leave-one-out scheme to evaluate the impact of each of the three aspects -- topic coherence, temporal dynamics, and historical comments -- on the performance of TGBully. 
We also include HAN for reference because TGBully will be converted to HAN when the modeling of the three aspects is removed simultaneously.

From Fig.~\ref{fig:ablation}, we observe that all three aspects are important to session-level cyberbullying detection.
In particular, the performance of TGBully exacerbates the most when topic coherence is removed from the learning process. Therefore, it is crucial to model semantic interaction between comments.
By explicitly connecting the edge weights to the modeling of topic coherence, TGBully enforces the correlations between semantically related comments meanwhile reduces the noise induced by unrelated comments. Modeling temporal dynamics also brings substantial performance improvement. Results show that transforming time intervals between comments into edge weights in the temporal graph can help capture the evolving characteristics of sessions.
Moreover, the consideration of historical comments also enhances the performance of TGBully, indicating the effectiveness of modeling users' personalities and language behavior for cyberbullying detection.
% Because not all historical comments are closely relate to current sessions, the improvement is less obvious when compared with the other two components.
By combining all three aspects together, TGBully can achieve the best performance overall.

\subsection{Influence of Data Imbalance}
Cyberbullying datasets are typically imbalanced. A common solution to extremely skewed data is data imputation, e.g., data oversampling. 
To further examine how the common data imputation techniques impact the model performance (\textbf{RQ3}), we artificially create a balanced dataset by oversampling data from the minority class. Specifically, 
We use SMOTE~\cite{chawla2002smote} to synthesize new bullying examples in the feature space and illustrate the performance comparisons of various methods in Table~\ref{oversample}.
We observe that oversampling brings converse effects to different methods.
It improves performances of LR, SVM, XGBoost, and LSTM, and decreases that of HAN, HANCD, and TGBully.
Two reasons may account for this.
First, methods such as LR, SVM, XGBoost, and LSTM only exploit semantic information of comments.
Since they capture information from a single modality, their classifiers may benefit from a more balanced data distribution in the feature space.
Second, other approaches (HAN, HANCD, and TGBully) exploit multi-model information, e.g., comment hierarchy, temporal dynamics , and user history.
Nevertheless, synthetic data assumes independence between different modalities and ignores their correlation.
Thus, they may add extra noise to models and harm their performances.

\begin{table}[t]
\centering
\caption{Performance comparison of different methods with oversampled dataset.}
\resizebox{0.47\textwidth}{!}{
\begin{tabular}{|c||ccc|ccc|}
\hline
             & \multicolumn{3}{c|}{\textbf{Oversampled}} & \multicolumn{3}{c|}{\textbf{Original}} \\ \hline
             & \small{\textbf{Recall}}  &
             \textbf{F1} & 
             \textbf{AUC} & 
             \small{\textbf{Recall}}     &
             {\textbf{F1}} &
             \textbf{AUC} 
             \\ \hline
LR   &\textbf{69.50}&\textbf{71.27}&\textbf{78.81}&59.06&69.12&76.97\\
SVM   &\textbf{60.99}&\textbf{69.08}&\textbf{77.06}&59.32&69.03&76.94\\
XGBoost   &\textbf{73.05} &\textbf{75.74}&\textbf{81.90}& 71.55&75.25&81.68\\
LSTM      &\textbf{70.44} &\textbf{68.75} &\textbf{85.25} &70.01&67.42 &83.87 \\
HAN       &77.93&76.75&\textbf{91.47}&\textbf{78.21}&\textbf{76.99}&91.16 \\
HANCD     &\textbf{77.36}&77.30&90.66 &76.91&\textbf{77.55}&\textbf{91.32}\\
TGBully    &79.02&78.67&91.81&\textbf{82.57} &\textbf{80.97} &\textbf{92.91} \\

\hline
\end{tabular}
}
\label{oversample}
\end{table}

\subsection{Case Study}
To answer \textbf{RQ4}, we investigate the capability of our model to detect bullying instances at the comment-level.
Here, we use the comment-level attention weights that indicate the relative importance of each user comment for classification. The underlying assumption is that comments with higher attention weights are more likely to be bullying instances.
To validate this, we ask three graduate students (two from computer science and one from psychology) to manually annotate the user comments in a total number of 100 sessions.
The final label of each comment is determined by a majority vote.
Among the 100 sessions with comment-level labels, we randomly select a cyberbullying session and a non-bullying session for illustration.
We aim to visually testify the effectiveness of our method and answer \textbf{RQ4}. The results can be seen in Fig.~\ref{fig:casestu}.
In addition to the user-level attention weights, we also present graph weights and word-level attention weights in TGBully. 

From Fig.~\ref{fig:casestu}, we make the following observations:
(1) TGBully can effectively recognize bullying users (comments) based on the learned attention weights.
As illustrated in Fig.~\ref{fig:bully}, as the $2^{nd}$, $4^{th}$ and $5^{th}$ comments contain humiliating words, they are predicted with higher weights indicating a high probability of being bullying comments. 
We found that the results are lined with ground-truth labels.
In contrast, comments in Fig.~\ref{fig:nonbully} are non-bullying and their attention weights are more evenly distributed.
(2) TGBully encodes temporal dynamics and topic coherence between comments via the GAT.
Comparing Fig.~\ref{fig:bully} and Fig.~\ref{fig:nonbully}, we find that graph weights of bullying sessions are relatively dense and concentrate on specific nodes.
For instance, the $2^{nd}$, $4^{th}$ and $5^{th}$ comments convey similar semantic meaning.
Accordingly, their interactions are highlighted by the graph weights.
Meanwhile, apart from being coherent in topic, the $4^{th}$ and $5^{th}$ comments are also consecutive in time.
As a result, their correlations are more obvious than that between the $2^{nd}$ and $4^{th}$ comments.
By contrast, weights of the non-bullying session largely concentrate on diagonals, indicating that they tend to focus on self interactions of comments and reflect limited cross-comment interactions.
Despite the overlapped words and temporal proximity between comments, our method doesn't capture significant cross-comment interaction.
This may happen because the GAT model only aims to capture specific bully-related interaction.
(3) TGBully effectively selects offensive words. For instance, words such as ``terrible'', ``ni**a'', and ``bit*h'' in Fig.~\ref{fig:nonbully} have larger attention weights.
We also notice that TGBully recognizes some salient words in a non-bullying session, which indicates it is discriminative to bullying words.
\section{Conclusion and Future Work}
In this work, we make a novel observation that in a social media session, user interaction is embodied by the relations between their posted comments, and such interaction dynamically evolves with time.
We further show that the topic coherence and temporal dynamics in the modeling of user interactions jointly contribute to capturing the repetitive characteristic of cyberbullying behavior.
The core contribution of this work is the proposed TGBully, a temporal graph-based cyberbullying detection framework with a focus on user interaction. TGBully consists of three modules.
The first \textit{semantic context modeling} module hierarchically constructs user comments from sequences of words and infers users' language behavior and personality from their historical comments.
The second \textit{temporal graph interaction learning} module first constructs the user interaction graph. 
It then encodes the topic coherence and temporal proximity through a bully-featured GAT. 
The third \textit{session classification} module aggregates the information learned from user interaction in a session and performs the final session-level classification.
We empirically evaluate the effectiveness of our approach with the tasks of session-level bullying detection and comment-level case study.

Our work opens several key avenues for future work. One such direction is to combine the explicit reply/social network with the introduced implicit temporal graph to together model the user interaction in a social media session. Additionally, we can incorporate other important aspects of cyberbullying into the temporal interaction graph. For instance, we can consider different roles of users (e.g., bullies and victims) and study how the role of a user evolves over time in the graph modeling process. Further investigations are needed to apply TGBully to detecting comment-level cyberbullying instances at scale.
% Meanwhile, more exploration can be made towards better exploiting the social relationship between users, such as introducing social network structural information and combing it with the temporal graph.
\section{Ethics Statement}
The common understanding of the importance of data privacy and security is shared by the authors. To secure user privacy, we replaced usernames and ids (include their mentioning in other users' comments) with numerical indices before the annotating and prepossessing process. The insulting or offensive terms and the figures used in this paper are for illustration only, they do not represent the ethical attitude of the authors.
\section*{ACKNOWLEDGEMENTS}
This material is based upon work supported by the Office of Naval Research (ONR) Grant N00014-21-1-4002, Association for Research Libraries (ARL) Grant W911NF2020124, and National Science Foundation (NSF) Grant \#2036127.

\bibliographystyle{ACM-Reference-Format}
\balance 
\bibliography{sample-base}
\end{document}